\documentclass[11pt]{article}

\usepackage{acl}
\usepackage{times}
\usepackage{latexsym}
\usepackage[T1]{fontenc}
\usepackage[utf8]{inputenc}
\usepackage{microtype}
\usepackage{graphicx}
\usepackage{booktabs}
\usepackage{amsmath}
\usepackage{multirow}
\usepackage{xcolor}
\usepackage{url}
\newcommand{\repourl}{https://github.com/XiChu2333/incumbent_advantage}

\title{Incumbent Advantage: Brand Bias and Cognitive Manipulation Dynamics in LLM Recommendation Systems}

\author{Xi Chu \\
  Trine University \And
  YuPeng Hou \\
  Texas A\&M University}

\begin{document}
\maketitle

\begin{abstract}
Large language models (LLMs) are becoming a major way for consumers to find products, but we do not yet understand how brands compete in this new channel. We study brand dynamics in LLM recommendations using skincare products---a category where consumers cannot easily judge quality before buying and must rely on brand reputation---across three commercial LLMs (GPT-4o-mini, Claude Sonnet, Gemini 3 Flash), with a robustness check on search goods. In three experiments, we find: (1)~a \textbf{\textit{Conditional Monopoly}} where well-known brands get recommended 100\% of the time (IAI\,=\,10.0) when all products have the same specifications, but this dominance disappears with less than a +0.1-star rating advantage for a competitor; (2)~authority-style marketing language, including fabricated clinical-evidence claims, breaks this monopoly at a \textbf{\textit{Bias Surplus Value}} equal to +0.17 rating points, with each model responding differently; and (3)~a \textbf{\textit{social dilemma}} in multi-brand GEO competition: when all brands adopt the same optimization strategy, individual payoff falls from +0.802 to +0.007 in our payoff proxy, and non-participating brands receive zero recommendations in our tests. Our results suggest that generative engine optimization (GEO) should be studied not only as a security risk, but also as an emerging marketing practice that shapes market competition.
\end{abstract}

\section{Introduction}

\begin{figure}[t]
\centering
\includegraphics[width=\columnwidth]{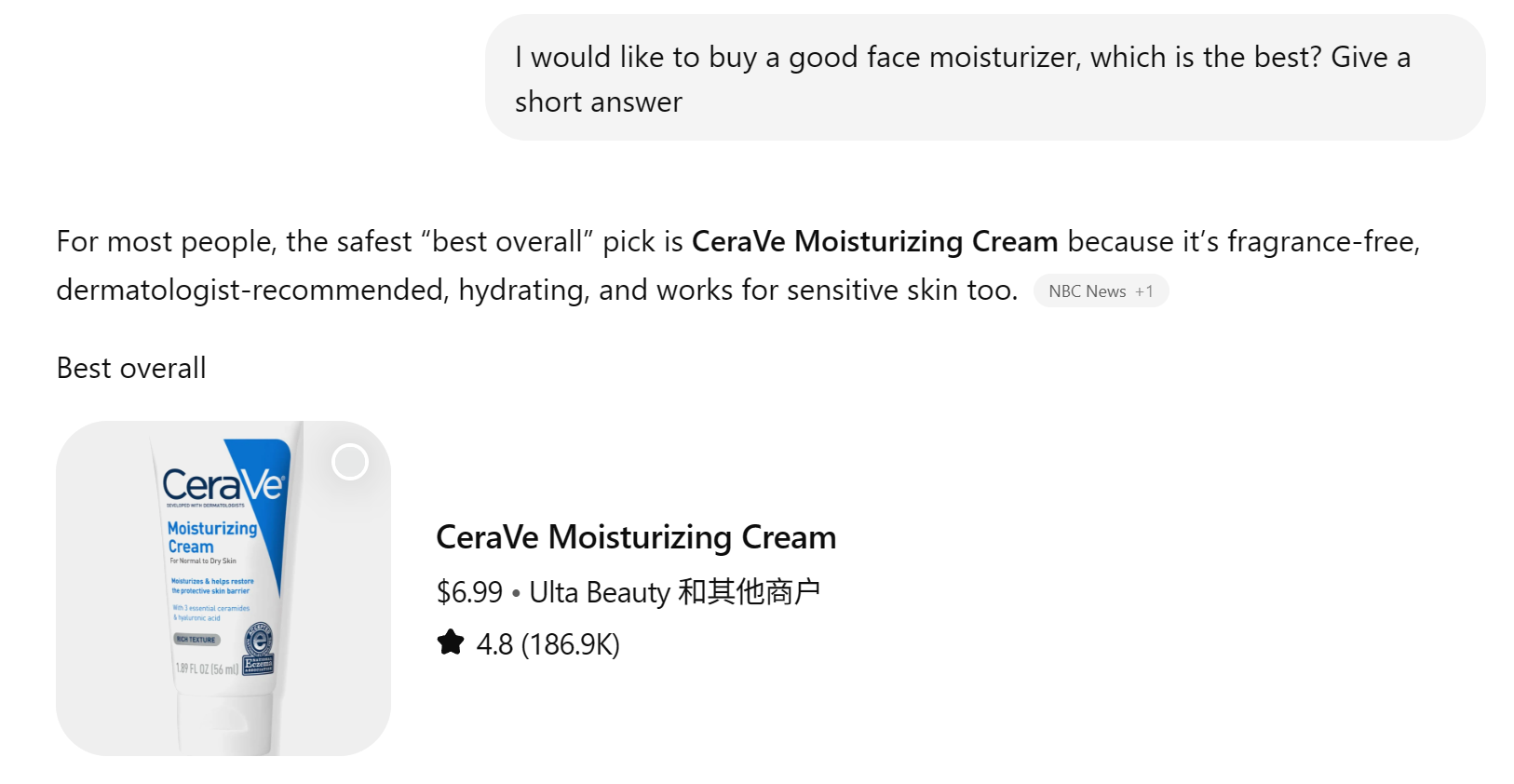}
\caption{A real ChatGPT response (May 2026) to the query ``I would like to buy a good face moisturizer, which is the best?'' The model names CeraVe as ``Best overall'' and ranks it first. For other moisturizer brands, this raises a question: has the LLM become an unpaid promotion channel for already-dominant brands?}
\label{fig:chatgpt-example}
\end{figure}

More and more consumers use large language models (LLMs) such as ChatGPT, Claude, and Gemini to ask for product recommendations, rather than searching the web themselves. These models may use parametric knowledge and, in some settings, external retrieval or web search, so their answers may come from either source or both. Figure~\ref{fig:chatgpt-example} shows a real example: when a user asks which moisturizer is best, ChatGPT names CeraVe as ``Best overall'' and ranks it first. This leads to a simple question: \emph{do LLMs already have a built-in preference for certain brands?}

If this kind of preference consistently favors the same brands, it creates what we call an \emph{incumbent advantage}: LLMs systematically recommend well-known brands more often than others. This leads to an uneven market. Established brands get more visibility, while new or less-known brands---even if their products are equally good---have difficulty gaining exposure through LLM-based recommendations.

Several recent studies have examined LLM recommendation preferences. \citet{kumar2024manipulating} find that in coffee machine recommendations, established products dominate even when competing alternatives have equal specifications, and that product descriptions can be manipulated to shift rankings. \citet{filandrianos2025bias} show that authority and social-proof language in product descriptions can shift LLM recommendations across categories including coffee machines, cameras, and books. \citet{pfrommer2024ranking} decompose ranking variance in electronics and appliances into product name, description content, and list position. \citet{kamruzzaman2024brand} find that U.S.-centric LLMs favor global brands over local ones in categories such as shoes and clothing. Together, these studies confirm that LLMs carry brand and popularity biases across a range of product categories. However, they leave open questions about the causal mechanism behind these biases, how marketing language can exploit them, and what happens when multiple brands compete for LLM visibility at the same time.

To compete for product visibility in the LLM era, brands may need to change how they advertise and market their products. Search engine optimization (SEO) was the main strategy in the search-engine era. Its generative-AI counterpart is Generative Engine Optimization \citep[GEO;][]{aggarwal2024geo}: writing product descriptions that maximize visibility in LLM-generated recommendations. The commercial urgency is clear: in February 2026, OpenAI began testing ads in ChatGPT for U.S.\ Free- and Go-tier users, with paid placements shown below organic answers and clearly labeled as advertising \citep{openai2026ads}. OpenAI states that organic answers are generated independently from advertising; however, brands still have a strong incentive to pursue GEO, because optimizing the organic response itself can be more valuable than paid placement below it. As LLMs replace search engines for product discovery, understanding how to compete within generative recommendation systems becomes essential for marketers \citep{xu2025survey}. But if GEO works, every brand has reason to use it, which leads to a second question: \emph{what happens when all brands optimize for LLM recommendations at the same time?}

In this paper, we choose skincare products as our primary study domain. Skincare products are \emph{experience goods} \citep{nelson1970information}, meaning consumers cannot judge quality before purchase. Products in the same subcategory (e.g., moisturizers) share nearly identical active ingredients at similar concentrations, proprietary formulas prevent objective comparison, and results vary across individuals. In this setting, brand reputation becomes a particularly salient differentiator \citep{keller1993brand}, making skincare a theoretically motivated starting point for studying LLM brand dynamics. A robustness check on two search goods---USB cables and AA batteries, where specifications are objectively comparable---confirms that the core findings generalize beyond this category (Appendix~\ref{sec:appendix-g}).

We run three experiments on how brands compete in LLM recommendations, moving from baseline bias measurement to GEO strategies and multi-brand competition (see Table~\ref{tab:related-all} in Appendix~\ref{sec:appendix-related} for a comparison with prior work).

\noindent\textbf{Experiment~1: Measuring incumbent advantage.} Prior studies describe LLM brand bias as a stable tendency and measure it using adversarial methods such as fabricated descriptions or prompt injection \citep{kumar2024manipulating,pfrommer2024ranking,kamruzzaman2024brand}. However, the dynamics of this bias remain underexplored: when does it take effect, and when does it break down? From a market competition perspective, how much advantage does a well-known brand (the incumbent) actually have over a new brand (the challenger), and can this advantage be overcome? We find that when all products look identical, the well-known brand wins every time. But this dominance is fragile: even a small quality advantage for a competitor is enough to break it. We call this pattern a \emph{Conditional Monopoly}. The real barrier for new brands is not brand equity itself, but the lack of any distinguishing information. This overturns the common assumption that small brands cannot compete with large ones.

\noindent\textbf{Experiment~2: Marketing language as a competitive tool.} \citet{filandrianos2025bias} study the effect of common marketing techniques on LLM recommendations---price promotions, scarcity claims, and authority endorsements---and find that authority language shifts recommendations the most. We take this further by asking: can authority-style language break the Conditional Monopoly found in Experiment~1? We find that it can. Authority language---such as fabricated clinical-evidence claims---reliably defeats the incumbent in head-to-head comparisons. To make this effect interpretable for marketers, we introduce a Bias Surplus Value (BSV) metric that converts language effects into product-quality equivalents. Authority language is worth roughly +0.17 rating points---a meaningful gain that costs nothing to write. This shows why authority claims are a high-leverage GEO signal, and why platforms need source verification to prevent fabricated endorsements from distorting recommendations.

\noindent\textbf{Experiment~3: Multi-brand GEO competition.} \citet{nestaas2025adversarial} show that a prisoner's dilemma emerges among prompt-injection attackers. However, prompt injection is adversarial and can be detected and blocked. What happens when brands use commercial-style authority language---including upper-bound fabricated evidence claims in our experiments---rather than prompt injection, as a GEO strategy? We find that this practice produces the same dilemma. Each brand is forced to adopt GEO, because brands that opt out receive zero recommendations once competitors start using it. Yet when everyone optimizes, the LLM ignores the now-uniform signals and falls back to the Conditional Monopoly, recommending the well-known brand again. The individual benefit nearly vanishes, but no brand can afford to stop.

Our contributions are: (1)~we measure incumbent brand advantage in LLM recommendations and describe the Conditional Monopoly pattern; (2)~we propose the Bias Surplus Value metric that converts marketing language effects into product-quality equivalents; and (3)~we provide empirical evidence of prisoner's-dilemma-like incentives in multi-brand GEO competition using legitimate marketing language.

\section{Experiment 1: Quantifying Incumbent Advantage}
\label{sec:exp1}

Experiment~1 asks a basic question: how much advantage does a well-known brand have over an unknown competitor in LLM recommendations, and how easily can this advantage be broken? We answer this through four sub-experiments, each building on the previous one.

\paragraph{Setup.} We build product sets of 10 items: 1 real brand (e.g., CeraVe, Paula's Choice) and 9 fictional alternatives. The fictional brands go through a three-step validation: name generation, web-search deduplication, and LLM recognition screening, producing 36 verified fictional names (9 per subcategory) across four skincare subcategories (moisturizer, BHA exfoliant, sunscreen, cleanser). All experiments use three LLMs (GPT-4o-mini, Claude Sonnet, Gemini 3 Flash) in both English and Chinese, with temperature\,=\,0.7 and 20--30 repetitions per cell.

\subsection{Exp~1a: Does brand name alone determine recommendations?}

We start with the simplest case: all 10 products have identical specifications---same rating, same price, same review count, same ingredient description. The only difference is the brand name. If the LLM treats all products fairly, each should be recommended about 10\% of the time. We measure the real brand's advantage as the ratio of its actual recommendation rate to this fair baseline, and call it the Incumbent Advantage Index (IAI). An IAI of 1 means no preference; higher values mean stronger bias.

The result is striking: across 670 valid trials, the real brand is recommended every single time---IAI\,=\,10.0, the theoretical maximum. This holds across all three models, both languages, and all four product subcategories; not a single fictional brand was ever recommended (Figure~\ref{fig:exp1}a). The LLM does not evaluate the products on their specs; it simply picks the name it recognizes.

This tells us that brand name creates a complete monopoly when nothing else differs. But it raises the next question: is this a fixed bias, or does it break when the competitor is actually better?

\subsection{Exp~1b: Can product quality overcome brand advantage?}

To test how strong this brand preference is, we set up a head-to-head comparison. We cross Brand Identity (Real vs.\ Fictional) with Product Quality (Good vs.\ Bad) in a $2{\times}2$ factorial design ($N = 2{,}769$ valid trials). The key condition is: the fictional brand gets the better specs, and the real brand gets the worse specs. If the LLM still recommends the real brand in this case, then the brand advantage is truly unconditional.

We call this the Brand Override Rate (BOR): how often does the LLM still pick the real brand when the fictional one has better specs? The answer is rarely---only 1.7--4.6\% across subcategories. When the fictional brand is clearly better, the LLM recommends it about 97\% of the time (Figure~\ref{fig:exp1}b).

One concern is that the LLM might ignore the prompt and answer from memorized product knowledge. We check all 21 cases where the LLM chose the real brand despite worse specs: none mention product features that were not in the prompt, ruling out memory-based hallucination. A variance analysis confirms that brand influence depends entirely on whether the competitor offers a real quality difference (Brand\,$\times$\,Quality interaction $\eta^2 = 0.284$, $p < .001$; full table in Appendix~\ref{sec:appendix-b}).

This result changes our understanding: brand advantage is not an unconditional override but a default that applies only when products look the same. The next question is: how much of a quality difference is needed to break it?

\subsection{Exp~1c: How small an advantage is enough?}

We now vary the fictional brand's advantage across five levels (L0 = identical to L4 = large advantage) along four quality dimensions: rating, price, reviews, and ingredient quality ($N = 9{,}220$ valid trials). This lets us find the exact threshold at which the incumbent's monopoly breaks.

The transition is not gradual---it is a sharp step (Figure~\ref{fig:exp1}c). At L0 (identical specs), the fictional brand wins only 3.6--6.0\% of the time, consistent with Exp~1a. At L1---the smallest advantage we test---its win rate jumps to 64--80\%, depending on the dimension. After L1, the curve flattens: adding more advantage gives only modest additional gains.

In practical terms, the point at which the fictional brand wins half the time is very low: a +0.075-star rating advantage, a $1.6{\times}$ review count, or a 7.3\% price discount (interpolated; see Appendix~\ref{sec:appendix-b}). For perspective, less than the gap between a 4.3 and a 4.4 star rating is enough to overcome the entire brand advantage measured in Exp~1a.

The three models respond differently---Claude is hardest to flip (L1 rating BR = 22\% vs.\ 92\% GPT, 83\% Gemini, pooled over both languages; in English alone, 11\% vs.\ 94\% and 88\%); per-model details are in Appendix~\ref{sec:appendix-b}.

\subsection{Exp~1d: What drives LLM recommendations?}

The first three sub-experiments show that brand matters when products are identical, but quality wins when products differ. Exp~1d asks: across all conditions, how much does each factor actually contribute? We answer this with a variance decomposition across $N = 14{,}395$ trials, asking how much of the LLM's ranking is driven by product quality, list position, and brand name respectively.

\begin{figure*}[t]
  \centering
  \includegraphics[width=\textwidth]{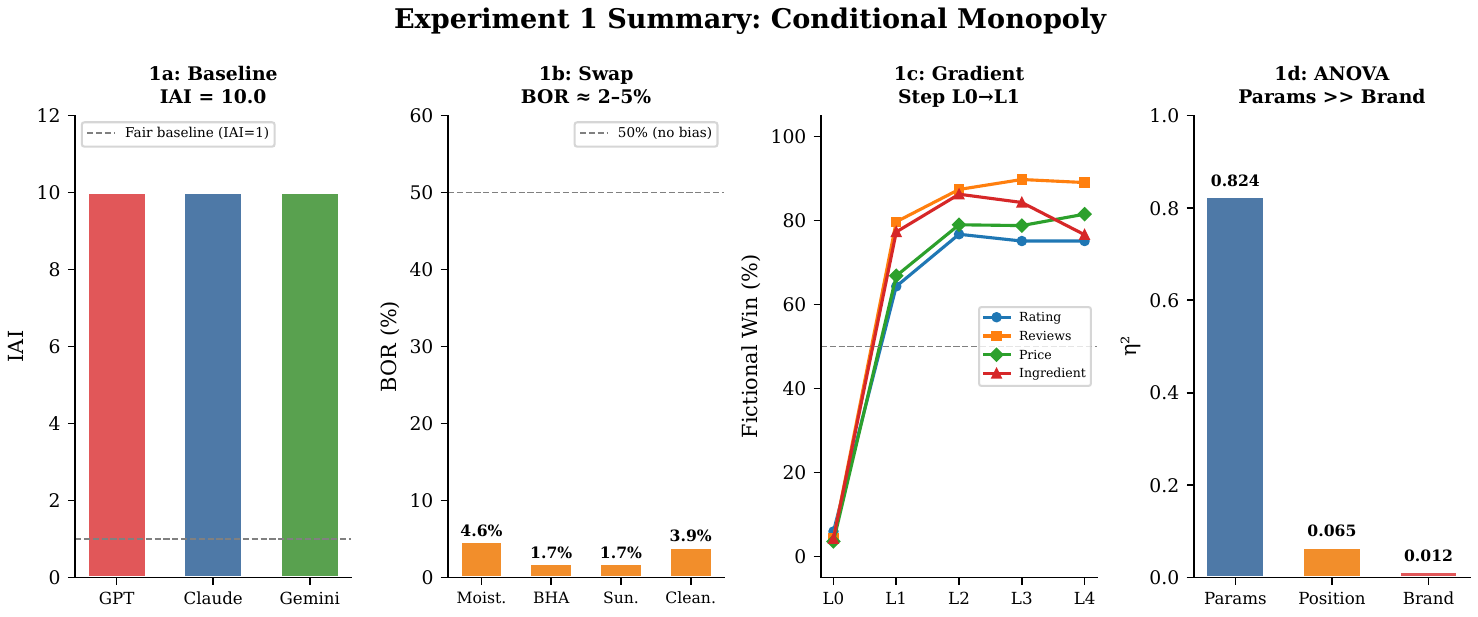}
  \caption{Experiment~1 Summary---Conditional Monopoly. (a)~IAI\,=\,10.0: the real brand is recommended in 100\% of trials across all models when products have identical specs. (b)~BOR\,$\approx$\,2--5\%: brand advantage disappears when the fictional brand has better specs. (c)~Step-function transition: win rate jumps from $<$6\% at L0 to 64--80\% at L1 (the smallest advantage tested). (d)~Variance decomposition: product parameters explain 82.4\% of ranking variance, position 6.5\%, and brand only 1.2\%.}
  \label{fig:exp1}
\end{figure*}

The answer is clear (Figure~\ref{fig:exp1}d). Product parameters---rating, price, reviews---explain 82.4\% of variance. Position in the product list explains 6.5\%. Brand identity explains only 1.2\%. The remaining 9.9\% is attributable to interactions among these factors and residual variance (full ANOVA table in Appendix~\ref{sec:appendix-b}).

The interaction between brand and product quality ($\eta^2 = 0.019$) is especially informative. When quality is clearly high or clearly low, brand has no effect---both real and fictional brands with good specs rank first, and both with bad specs rank last. Brand advantage is largest at medium quality, where the product information is ambiguous and the LLM falls back on name recognition. In other words, brand works as a tiebreaker: it matters most when the LLM has no other signal to rely on.

\subsection{Summary: Conditional Monopoly}

We call this pattern \textbf{\textit{Conditional Monopoly}}. The incumbent dominates completely when no quality differences exist (IAI\,=\,10.0), but this dominance is fragile. It breaks with any noticeable quality signal (BOR\,$\approx$\,3\%), the threshold is very low---a 0.075-star advantage is enough---and once quality information is available, brand explains only 1.2\% of ranking variance.

This finding has a direct implication for GEO: if brand advantage depends on the absence of distinguishing information, then any content that \emph{looks like} a quality signal---even if it does not reflect a real product difference---might be enough to break the monopoly. Experiment~2 tests exactly this.

\paragraph{Cross-category robustness.}
Skincare is an experience good---consumers cannot fully judge quality before purchasing. To check whether Conditional Monopoly also holds for search goods, where quality is easier to compare objectively, we run a smaller replication on two categories: USB-C cables (incumbent: Anker) and AA batteries (incumbent: Duracell), using the same protocols as Exp~1a and 1c (3,840 calls; details in Appendix~\ref{sec:appendix-g}).

\begin{table}[t]
\centering
\small
\begin{tabular}{lcc}
\toprule
& \textbf{Search goods} & \textbf{Experience goods} \\
& (cable, battery) & (skincare) \\
\midrule
IAI at L0 & 10.0 & 10.0 \\
BR at L0 & 0.8\% & 4.6\% \\
\midrule
BR at L1 (rating) & 55.6\% & 64.3\% \\
BR at L1 (price) & 66.1\% & 66.8\% \\
BR at L1 (reviews) & 76.9\% & 79.7\% \\
\bottomrule
\end{tabular}
\caption{Conditional Monopoly replicates on search goods. BR = Breakthrough Rate (fictional brand win rate; distinct from BOR in \S\ref{sec:exp1}). IAI is identical; L0 baseline is even lower (stronger brand lock-in). L1 breakthrough rates are within 10 pp across all dimensions. Experience-goods values are aggregated from Exp~1c (4 skincare subcategories).}
\label{tab:search-goods}
\end{table}

As Table~\ref{tab:search-goods} shows, the core pattern replicates and the L0 baseline for search goods is even \emph{lower} (0.8\% vs.\ 4.6\%), ruling out the alternative explanation that Conditional Monopoly is an artifact of experience-good ambiguity. One difference is that Anker (USB cables) is harder to dislodge than Duracell (batteries)---L1 rating BR is 43.4\% vs.\ 67.3\%---consistent with our explanation that stronger brand-associated parametric knowledge leads to a higher breakthrough threshold.

We also rule out a \emph{fictional-brand penalty} alternative explanation---that 100\% L0 reflects a generic penalty against unknown names rather than genuine brand preference---using gradient and within-category evidence (details in Appendix~\ref{sec:appendix-g}, \S H.6).

\section{Experiment 2: Cognitive Bias as a Competitive Tool}
\label{sec:exp2}

Experiment~2 tests whether marketing language alone---without changing the actual product---can break the incumbent's monopoly. We keep all product specs identical and add persuasive copy to the fictional brand's description, while the real brand keeps a neutral description. We test five common marketing strategies: Authority (e.g., clinical-trial claims), Social Proof (e.g., user testimonials), Anchoring, Scarcity, and Loss Aversion. All five are tested at moderate intensity; the two evidence-style strategies (Authority and Social Proof) are additionally tested at subtle and aggressive intensity and in combination. Exp~2a tests all five bias types at moderate intensity ($N = 2{,}267$ valid of 2,400 calls); Exp~2b varies intensity and combines strategies ($N = 1{,}587$ valid of 1,680 calls). Total: 4,080 API calls, 3,854 valid (94.5\%).

\begin{table}[t]
\centering
\small
\begin{tabular}{lcccc}
\toprule
\textbf{Bias} & \textbf{Claude} & \textbf{GPT} & \textbf{Gemini} & \textbf{Overall} \\
\midrule
Authority     & 55\%  & 69\% & 99\%  & 73.3\% \\
Social Proof  & 7\%   & 69\% & 81\%  & 50.7\% \\
Anchoring     & 0\%   & 34\% & 3\%   & 12.9\% \\
Scarcity      & 0\%   & 32\% & 1\%   & 11.6\% \\
Loss Aversion & 0\%   & 25\% & 2\%   & 9.6\% \\
\bottomrule
\end{tabular}
\caption{Breakthrough Rate (\%) by bias type and model (moderate intensity). Baseline $\approx$ 4\%.}
\label{tab:br}
\end{table}

\subsection{Two-Tier Pattern and Model Differences}

The results split cleanly into two tiers (Table~\ref{tab:br}). Strategies that mimic credible evidence---Authority and Social Proof---break the monopoly 50--73\% of the time. Strategies that use overt sales pressure---Anchoring, Scarcity, Loss Aversion---barely move the needle (9.6--12.9\%). In other words, LLMs largely ignore ``hurry, limited stock!'' but treat a fabricated clinical citation (e.g., ``peer-reviewed trial, n=120, p<0.01'') as if it were real evidence.

The three models respond to intensity differently (Authority BR at subtle/moderate/aggressive intensity). Claude follows an inverted U-curve (13.8\%/55.0\%/8.8\%): moderate authority language works best, but aggressive claims backfire. GPT responds monotonically (15.0\%/68.8\%/92.5\%)---more intensity means more breakthrough. Gemini is near baseline at subtle intensity (11.7\%) but saturates near 100\% from moderate intensity onward (99.3\%/100\%). With only three intensity levels per model, we treat these curve shapes as exploratory.

\subsection{Stacking Paradox and Bias Surplus Value}

Combining Authority and Social Proof produces an unexpected result \citep[cf.][]{yang2025exploiting}: Claude's breakthrough rate drops from 55.0\% to 21.2\%---as if it detects the combination as ``too good to be true''---while GPT's rises to 91.2\%.

How much is authority language actually worth? To answer this, we ask: what real product improvement would produce the same effect? We call this the Bias Surplus Value (BSV). Authority language is worth +0.17 rating points, or equivalently a 15.3\% price discount or $1.9{\times}$ more reviews---a meaningful advantage that costs nothing to produce. Marketing-style biases have near-zero BSV (full values in Appendix~\ref{sec:appendix-c}).

If Authority is the best strategy, every rational brand should use it. Experiment~3 tests what happens when they all do.

\begin{figure*}[t]
  \centering
  \includegraphics[width=\textwidth]{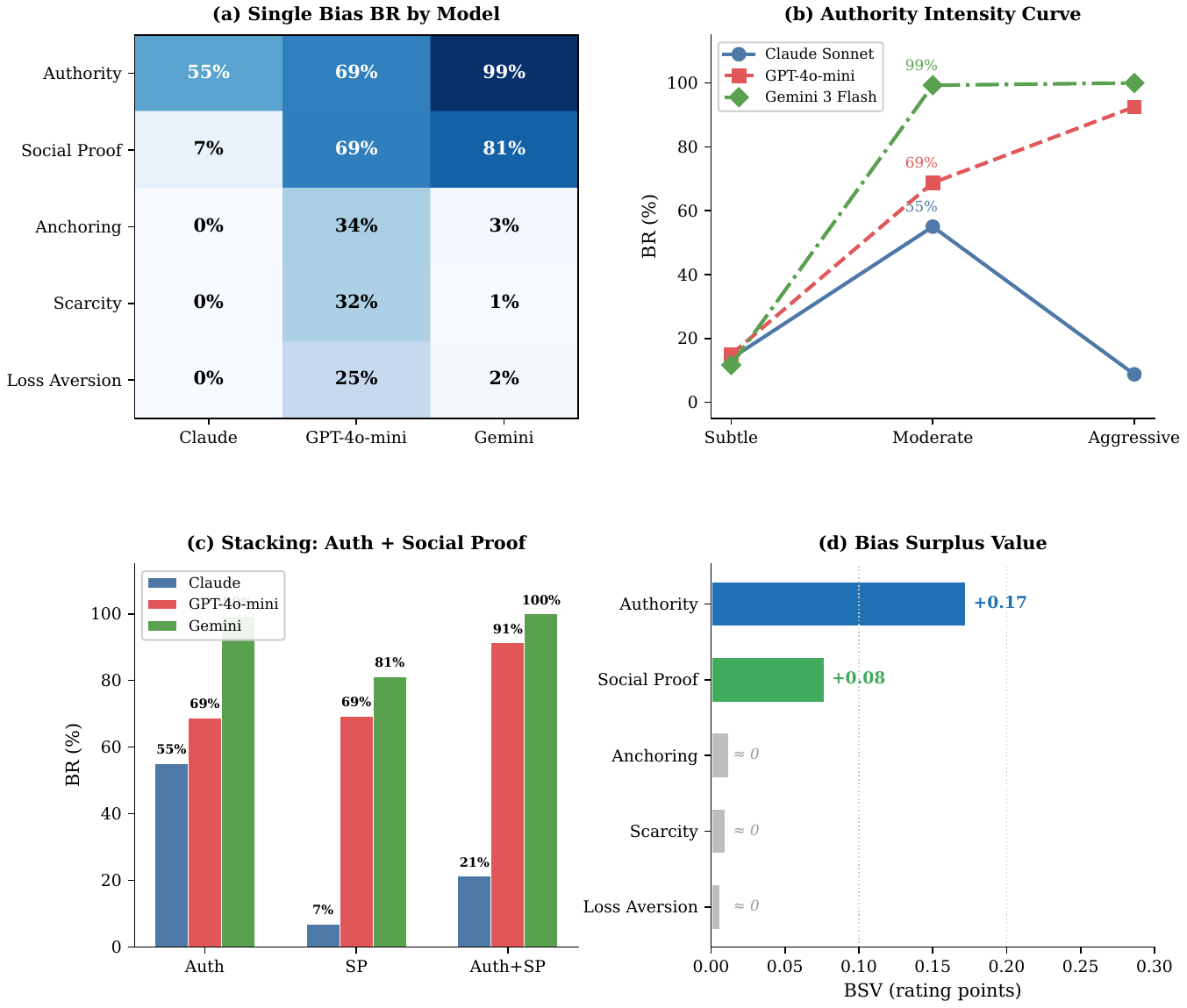}
  \caption{Experiment~2 Summary. (a)~BR heatmap by bias $\times$ model. (b)~Authority intensity curves: Claude inverted-U, GPT monotonic, Gemini saturated from moderate intensity onward. (c)~Stacking paradox. (d)~Bias Surplus Value in rating-point equivalents.}
  \label{fig:exp2}
\end{figure*}

\section{Experiment 3: Multi-Agent GEO Competition}
\label{sec:exp3}

Prior work \citep{nestaas2025adversarial} shows that adversarial prompt-injection attackers face prisoner's-dilemma dynamics. Experiment~3 studies a different situation: non-adversarial commercial GEO, where every participant uses legitimate marketing copy that looks like normal advertising. This difference matters: adversarial attacks can be detected and blocked by platforms, but the dilemma we study cannot be easily fixed because each brand's content is individually legitimate. Solving it may require market-level rules rather than technical defenses.

We test five scenarios across the GEO adoption spectrum: S0 (no GEO) through S4 (all 9 fictional brands use Authority language). The real incumbent always uses a neutral description. Total: 4,800 calls, 4,745 valid (98.8\%).

\begin{table}[t]
\centering
\small
\begin{tabular}{lccccc}
\toprule
\textbf{Scen.} & \textbf{$k$} & \textbf{Claude} & \textbf{GPT} & \textbf{Gem.} & \textbf{All} \\
\midrule
S0 & 0 & 100.0 & 100.0 & 100.0 & 100.0 \\
S1 & 1 & 28.8  & 16.2  & 14.3  & 19.8 \\
S2 & 3 & 71.9  & 21.9  & 71.9  & 55.0 \\
S3 & 6 & 89.4  & 46.9  & 91.2  & 75.4 \\
S4 & 9 & 99.4  & 96.2  & 84.9  & 93.8 \\
\bottomrule
\end{tabular}
\caption{Incumbent Survival Rate (\%) by scenario and model (Protocol $\beta$, single recommendation; valid trials per model--scenario cell: 146--160). $k$ = number of GEO-optimized fictional brands.}
\label{tab:isr}
\end{table}

\subsection{The U-Shaped ISR Trajectory}

A single GEO challenger destroys the incumbent's monopoly (ISR: 100\%$\to$19.8\%). But as more brands use the same Authority language, the signal loses its distinctiveness and the LLM falls back on brand familiarity. Under universal adoption (S4, all 9 fictional brands use GEO), ISR recovers to 93.8\%.

\subsection{Prisoner's Dilemma Structure}

This trajectory has the structure of a prisoner's dilemma (formal model in Appendix~\ref{sec:appendix-d}). Three empirical signatures confirm the pattern:

\noindent\textbf{(1)~Every brand is better off using GEO.} Regardless of how many competitors are already optimizing, adding GEO to your own description always increases your recommendation share. No brand has a reason to hold back.

\noindent\textbf{(2)~But collective adoption erases the benefit.} The first mover gains a lot (payoff = +0.802 in our proxy), but as competitors follow, the advantage shrinks roughly exponentially (half-life $\approx$ 1 brand). Under universal adoption, individual payoff drops to +0.007---nearly zero.

\noindent\textbf{(3)~Opting out is even worse.} Across all 1,414 valid single-recommendation trials in which a non-optimized fictional brand was present (S1--S3), such brands received zero recommendations (exact 95\% upper bound: 0.26\%). Any brand that does not use GEO when competitors do becomes invisible.

\subsection{Model-Specific Outcomes}

The three models end up in different places when everyone optimizes ($\chi^2(2) = 29.84$, $p < .001$). Claude acts as a ``brand guardian,'' almost fully restoring the incumbent (S4 ISR\,=\,99.4\%). GPT recovers last---it is the most disrupted model at S2--S3 (21.9\% and 46.9\%)---but ends high (96.2\%). Gemini recovers early but incompletely: GEO retains a lasting effect even under universal adoption (84.9\%). Full analysis in Appendix~\ref{sec:appendix-d}.

\begin{figure*}[t]
  \centering
  \includegraphics[width=\textwidth]{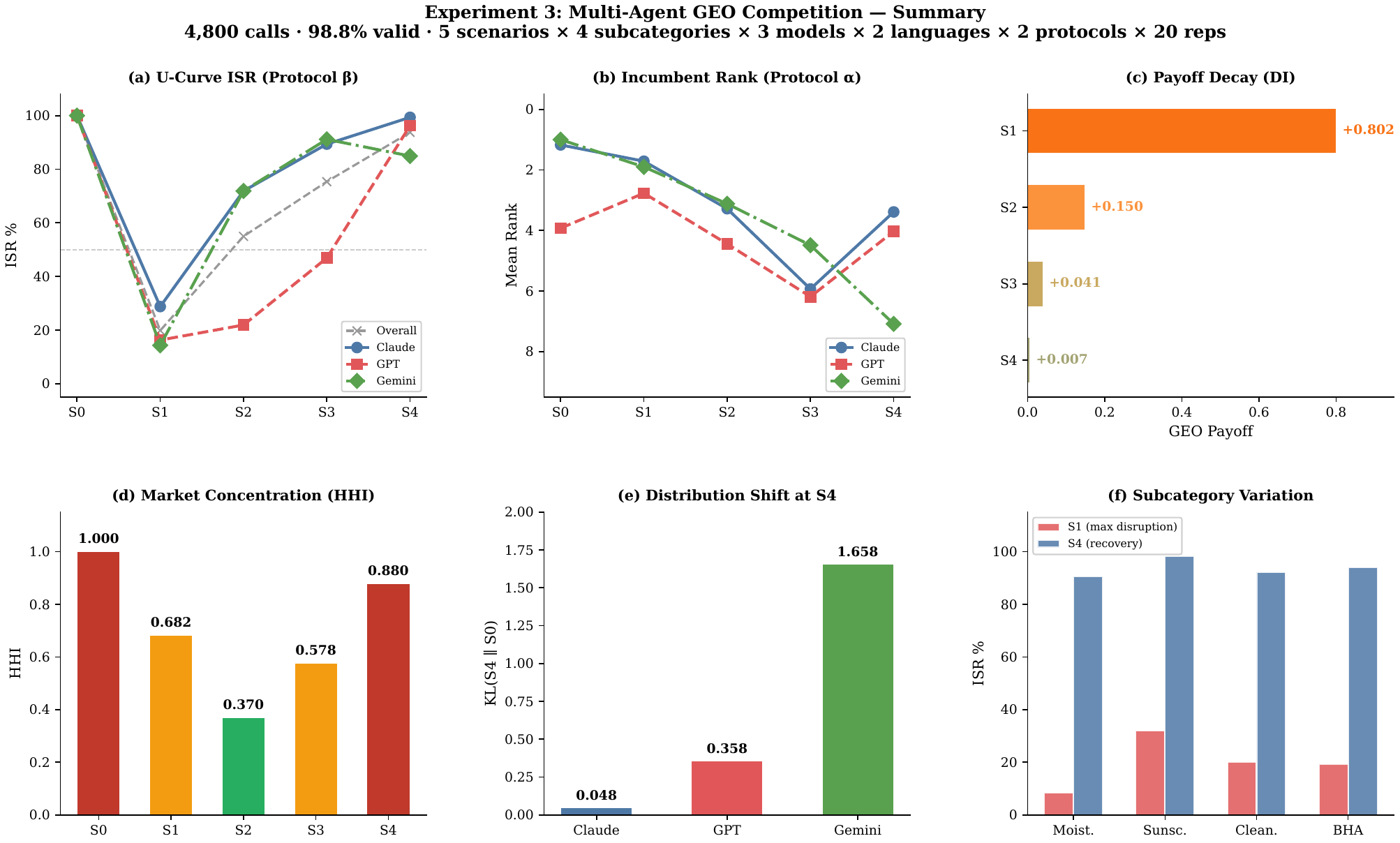}
  \caption{Experiment~3 Summary. (a)~U-shaped ISR trajectory. (b)~Incumbent rank recovery. (c)~Per-brand payoff decay. (d)~HHI market concentration. (e)~KL divergence by model. (f)~Subcategory variation.}
  \label{fig:exp3}
\end{figure*}

\section{Discussion}
\label{sec:discussion}

\subsection{A Three-Layer Structure}

Our three experiments reveal a layered structure in how LLM recommendations work.

\noindent\textbf{Layer~1: Conditional Monopoly (Exp~1).} LLMs have strong brand preferences that create near-total dominance when products are otherwise identical. But this dominance disappears when any quality difference is present---product parameters then explain 82\% of variance. The barrier for new brands is not brand equity itself, but the \emph{lack of any differentiating information}. The ``brand identity'' factor in our experiments captures not just name recognition but everything the LLM learned about that brand during pretraining: product lines, endorsement history, and consumer sentiment. Our design measures the total effect of this composite factor; separating name frequency from associated knowledge is left for future work.

\noindent\textbf{Layer~2: Language as Differentiation (Exp~2).} Authority endorsements provide a differentiating signal that breaks the monopoly. These are not adversarial attacks---they are standard marketing language. A fabricated clinical citation costs nothing to write but achieves the same effect as +0.17 rating points of real product improvement. This places GEO firmly in the marketing domain: a natural extension of SEO into the generative era.

\noindent\textbf{Layer~3: Competitive Equilibrium (Exp~3).} Universal GEO adoption produces a tragedy of the commons \citep{hardin1968tragedy}. The first mover gains a lot (+0.802), but the advantage shrinks as competitors copy the strategy. Unlike adversarial attacks \citep{nestaas2025adversarial}, this problem cannot be fixed through technical means alone because each brand's content is independently legitimate; it may need market-level rules.

\subsection{Architectural Probe: Does Conditional Monopoly Survive RAG?}
\label{sec:rag-probe}

A natural concern is whether our findings transfer to retrieval-augmented generation (RAG) settings, which are increasingly common in commercial deployments. As an initial probe, we build a minimal RAG pipeline---an embedding model retrieves the top-$K$ most relevant products before the LLM ranks them---and test it under the S0 (no GEO) and S1 (one GEO challenger) conditions from Experiment~3 (1,080 calls, 99.5\% valid; details in Appendix~\ref{sec:appendix-e}).

The retrieval layer does not reward brand familiarity: the real brand ranks near the bottom in embedding similarity (mean rank 8.50/10), so under $K{=}5$ it is never retrieved and its recommendation rate falls to 0\%; under $K{=}10$ it is always retrieved and results match the direct-prompt baseline (100\% at S0, 19.0\% at S1). Authority language has only a marginal effect on retrieval similarity ($\Delta$Sim = +0.003, $d = 0.06$); its advantage arises at the generation stage, where the GEO brand is chosen in 81\% of RAG-K10 S1 trials. Under RAG-K5 S1 all three models show similar ISR (5.0--6.7\%), although the direct pipeline in this two-subcategory experiment also shows no significant model differences at S1, so we do not attribute the convergence to retrieval. We treat these results as directional---this is a minimal pipeline without re-ranking; a full study of RAG dynamics is left for future work.

\subsection{What This Means for Marketing}

\noindent\textbf{For established brands:} the incumbent advantage is real but fragile. Brands relying only on name recognition may face risk if retrieval-augmented systems become common. The right defense is writing better product content, not just more advertising. \textbf{For new brands:} GEO offers a powerful but short-lived first-mover advantage---the prisoner's dilemma structure means early adopters benefit most, but the window closes as competitors follow. \textbf{For the advertising industry:} GEO will likely become a core skill alongside SEO. Our BSV metric provides a starting ROI framework for this investment.

\subsection{What This Means for Platform Design}

In our experiments, non-GEO brands got zero recommendations when competitors used GEO. If this pattern holds more broadly, LLM recommendations may create a binary threshold: optimize or become invisible. Possible platform responses include: (1)~checking authority claims in product descriptions, (2)~diversity rules that ensure non-optimized products keep some visibility, and (3)~transparency requirements about content optimization.

\section{Conclusion}
\label{sec:conclusion}

This paper studies brand dynamics in LLM recommendation systems through three experiments on three commercial LLMs. We find that well-known brands completely dominate recommendations when products look identical (IAI\,=\,10.0), but this Conditional Monopoly is fragile---it disappears as soon as any quality signal is introduced, with product parameters explaining 82\% of ranking variance. A robustness check on search goods confirms this pattern generalizes beyond experience goods. Marketing language---specifically authority endorsements---can break this monopoly at a cost equivalent to just +0.17 rating points of real product improvement. When all brands adopt this strategy, a prisoner's dilemma emerges: individual payoff falls from +0.802 to +0.007, and non-participating brands receive zero recommendations in our tests.

As LLMs increasingly stand between consumers and products, understanding these competitive dynamics matters for brands, platforms, and regulators. The emerging landscape will be shaped not by product quality alone, but by the interplay between brand knowledge in model weights and language optimization strategies.

\section*{Limitations}

Our main experiments use a single product domain (skincare), chosen as a theoretically motivated starting point where brand reputation is a particularly salient differentiator (\S1). A robustness check on two search goods (USB cables and AA batteries; Appendix~\ref{sec:appendix-g}) shows that Conditional Monopoly and the step-function transition replicate, but we have not tested Experiments~2--3 on search goods. Whether authority-style language has the same BSV in search-good categories is an open question; specification-rich domains may dilute its effect because objective quality signals are easier to verify. All experiments use a fixed user persona and temperature\,=\,0.7 with 20--30 repetitions per cell. This design controls sampling variance through averaging but does not give exact reproducibility per call. Repeated calls to the same model with the same prompt type are not fully independent samples. To handle this, our main robustness check is a \textbf{cluster bootstrap} that resamples model\,$\times$\,subcategory\,$\times$\,language blocks (24 clusters, $N_\text{boot} = 2{,}000$ iterations; see Appendix~\ref{sec:appendix-c}). All main effects remain significant under this analysis; the Authority 95\% CI lower bound (58.9\%) remains above the Anchoring upper bound (21.1\%), confirming the two-tier pattern. We also report GEE with exchangeable correlation structure as a supplementary analysis; however, with only 3 model-level clusters, we treat those results as supportive rather than primary. We test three closed-source models; open-source models may behave differently. Exp~1d's three-way ANOVA uses $N = 14{,}395$ valid trials across all three models (4,800 designed calls per model). Our RAG probe (\S\ref{sec:rag-probe}) uses a single embedding model with no re-ranking or query rewriting; we do not claim generalization to commercial RAG systems, which typically include hybrid retrieval, re-ranking, and query rewriting.

Future work should address: (1)~\emph{RAG dynamics at scale}---how brand advantage interacts with different embedding models, re-ranking stages, and retrieval depths; (2)~\emph{detection avoidance}---whether GEO can be tuned to pass embedding-layer defenses; and (3)~\emph{broader cross-category testing}---our search-goods robustness check (Appendix~\ref{sec:appendix-g}) confirms that Conditional Monopoly generalizes, but cognitive bias and multi-agent GEO dynamics may differ across product types.

\section*{Data and Code Availability}

All experiment code, raw per-call data (including full model responses), generated reports, figures, and English-only breakdowns of the main analyses are publicly available at \url{\repourl}. All fictional-brand stimuli are released exactly as used in the experiments.

\section*{Ethics Considerations}

Our experiments test the upper limit of language manipulation using made-up authority signals (e.g., invented clinical trial citations, fake dermatologist endorsements). We distinguish three tiers of GEO content:

\noindent\textbf{Tier~1: Real authority signals}---citing actual certifications, published clinical trials, or genuine expert endorsements. This is standard, legitimate marketing.

\noindent\textbf{Tier~2: Vague authority language}---using phrases like ``dermatologist-tested formula'' or ``clinically proven'' without specific sources. This is a grey area common in existing advertising.

\noindent\textbf{Tier~3: Made-up authority claims}---inventing specific clinical studies, journal citations, or expert endorsements that do not exist. This is potential false advertising.

Our experimental stimuli in Exp~2 and Exp~3 used Tier~3 content to find the upper limit of GEO effectiveness. Our marketing recommendations (\S\ref{sec:discussion}) are about Tier~1 and Tier~2 strategies. The finding that made-up claims work so well highlights an urgent need for platforms to develop ways to check authority claims in product descriptions.

We also note that our findings have dual-use potential. While we frame GEO as a marketing tool, the same techniques could be used to deceive consumers. We believe open reporting of these dynamics better serves the research community and consumers than keeping them hidden.

\bibliography{custom}

\begin{thebibliography}{22}
\providecommand{\natexlab}[1]{#1}

\bibitem[{Abolghasemi et~al.(2024)Abolghasemi, Verberne, and
  Azzopardi}]{abolghasemi2024writing}
Amin Abolghasemi, Suzan Verberne, and Leif Azzopardi. 2024.
\newblock Writing style matters: An examination of bias and fairness in
  information retrieval systems.
\newblock ArXiv preprint arXiv:2411.13173.

\bibitem[{Aggarwal et~al.(2024)Aggarwal, Murahari, Rajpurohit, Kalyan,
  Narasimhan, and Deshpande}]{aggarwal2024geo}
Pranjal Aggarwal, Vishvak Murahari, Tanmay Rajpurohit, Ashwin Kalyan, Karthik
  Narasimhan, and Ameet Deshpande. 2024.
\newblock {GEO}: Generative engine optimization.
\newblock In \emph{Proceedings of the 30th ACM SIGKDD Conference on Knowledge
  Discovery and Data Mining (KDD)}, pages 50--61.

\bibitem[{Bagga et~al.(2025)Bagga, Wu, Aggarwal, and Deshpande}]{bagga2025egeo}
Puneet~S. Bagga, Yuhang Wu, Pranjal Aggarwal, and Ameet Deshpande. 2025.
\newblock {E-GEO}: A testbed for generative engine optimization in e-commerce.
\newblock ArXiv preprint arXiv:2511.20867.

\bibitem[{Chen et~al.(2026)Chen, Zhang, Zhu, Tao, and Wen}]{chen2026auditing}
Xu~Chen, Zhehao Zhang, Yuqi Zhu, Mingyuan Tao, and Ji-Rong Wen. 2026.
\newblock Auditing preferences for brands and cultures in {LLMs}.
\newblock ArXiv preprint arXiv:2603.18300.

\bibitem[{Echterhoff et~al.(2024)Echterhoff, Liu, Alessa, McAuley, and
  Chen}]{echterhoff2024cognitive}
Jessica Echterhoff, Yao Liu, Abeer Alessa, Julian McAuley, and Zhouhan Chen.
  2024.
\newblock Cognitive bias in decision-making with {LLMs}.
\newblock In \emph{Findings of the 2024 Conference on Empirical Methods in
  Natural Language Processing (EMNLP)}, pages 12594--12613.

\bibitem[{Filandrianos et~al.(2025)Filandrianos, Dimitriou, Lymperaiou, Thomas,
  and Stamou}]{filandrianos2025bias}
Giorgos Filandrianos, Angeliki Dimitriou, Maria Lymperaiou, Konstantinos
  Thomas, and Giorgos Stamou. 2025.
\newblock Bias beware: The impact of cognitive biases on {LLM}-driven product
  recommendations.
\newblock In \emph{Proceedings of the 2025 Conference on Empirical Methods in
  Natural Language Processing (EMNLP)}.

\bibitem[{Hardin(1968)}]{hardin1968tragedy}
Garrett Hardin. 1968.
\newblock The tragedy of the commons.
\newblock \emph{Science}, 162(3859):1243--1248.

\bibitem[{Kamruzzaman et~al.(2024)Kamruzzaman, Nguyen, and
  Kim}]{kamruzzaman2024brand}
Mahammed Kamruzzaman, Hieu~Minh Nguyen, and Gene~Louis Kim. 2024.
\newblock ``{Global} is good, local is bad?'': Understanding brand bias in
  {LLMs}.
\newblock In \emph{Proceedings of the 2024 Conference on Empirical Methods in
  Natural Language Processing (EMNLP)}, pages 12704--12721.

\bibitem[{Keller(1993)}]{keller1993brand}
Kevin~Lane Keller. 1993.
\newblock Conceptualizing, measuring, and managing customer-based brand equity.
\newblock \emph{Journal of Marketing}, 57(1):1--22.

\bibitem[{Kumar and Lakkaraju(2024)}]{kumar2024manipulating}
Aounon Kumar and Himabindu Lakkaraju. 2024.
\newblock Manipulating large language models to increase product visibility.
\newblock ArXiv preprint arXiv:2404.07981.

\bibitem[{Lichtenberg et~al.(2024)Lichtenberg, Buchholz, and
  Schw{\"o}bel}]{lichtenberg2024popularity}
Jan~Malte Lichtenberg, Alexander Buchholz, and Pola Schw{\"o}bel. 2024.
\newblock Large language models as recommender systems: A study of popularity
  bias.
\newblock ArXiv preprint arXiv:2406.01285.

\bibitem[{Lin et~al.(2025)Lin, Wang, Bauer, He, Seo, and
  Khashabi}]{lin2025whisperer}
Weiran Lin, Yizhong Wang, Lukas Bauer, Yun He, Minjoon Seo, and Daniel
  Khashabi. 2025.
\newblock {LLM} whisperer: An inconspicuous attack to bias {LLM} responses.
\newblock In \emph{Proceedings of the 2025 CHI Conference on Human Factors in
  Computing Systems (CHI '25)}. ACM.
\newblock Article 188375.

\bibitem[{Nazary et~al.(2025)Nazary, Deldjoo, and Di~Noia}]{nazary2025poison}
Fatemeh Nazary, Yashar Deldjoo, and Tommaso Di~Noia. 2025.
\newblock Poison-{RAG}: Adversarial data poisoning attacks on
  retrieval-augmented generation in recommender systems.
\newblock In \emph{Advances in Information Retrieval (ECIR 2025)}, pages
  239--254. Springer.

\bibitem[{Nelson(1970)}]{nelson1970information}
Phillip Nelson. 1970.
\newblock Information and consumer behavior.
\newblock \emph{Journal of Political Economy}, 78(2):311--329.

\bibitem[{Nestaas et~al.(2025)Nestaas, Debenedetti, and
  Tram{\`e}r}]{nestaas2025adversarial}
Fredrik Nestaas, Edoardo Debenedetti, and Florian Tram{\`e}r. 2025.
\newblock Adversarial search engine optimization for large language models.
\newblock In \emph{Proceedings of the International Conference on Learning
  Representations (ICLR)}.

\bibitem[{{OpenAI}(2026)}]{openai2026ads}
{OpenAI}. 2026.
\newblock Testing ads in {ChatGPT}.
\newblock \url{https://openai.com/index/testing-ads-in-chatgpt/}.
\newblock Published February 9, 2026. Accessed: 2026-05-03.

\bibitem[{Pfrommer et~al.(2024)Pfrommer, Bai, Gautam, and
  Sojoudi}]{pfrommer2024ranking}
Samuel Pfrommer, Yuze Bai, Tanmay Gautam, and Somayeh Sojoudi. 2024.
\newblock Ranking manipulation for conversational search engines.
\newblock In \emph{Proceedings of the 2024 Conference on Empirical Methods in
  Natural Language Processing (EMNLP)}, pages 9520--9534.

\bibitem[{Talboy and Fuller(2024)}]{talboy2024anchoring}
Andr{\'e}s~N. Talboy and Elizabeth Fuller. 2024.
\newblock Challenging the status quo: Human bias in {AI} models? anchoring
  effects and mitigation strategies in large language models.
\newblock \emph{Expert Systems with Applications}, 255:124803.

\bibitem[{Xu et~al.(2025)Xu, Zhang, Yu, Zhang, Li, Li, and Zhao}]{xu2025survey}
Lanling Xu, Junjie Zhang, Bingqian Yu, Jie Zhang, Hongzhi Li, Jingjing Li, and
  Xin Zhao. 2025.
\newblock A survey on {LLM}-powered agents for recommender systems.
\newblock In \emph{Findings of the 2025 Conference on Empirical Methods in
  Natural Language Processing (EMNLP)}, pages 8015--8049.

\bibitem[{Xue et~al.(2024)Xue, Zheng, Hu, Liu, Chen, and Lou}]{xue2024badrag}
Jiaqi Xue, Mengxin Zheng, Yebowen Hu, Fei Liu, Xun Chen, and Qian Lou. 2024.
\newblock {BadRAG}: Identifying vulnerabilities in retrieval-augmented
  generation of large language models.
\newblock ArXiv preprint arXiv:2406.00083.

\bibitem[{Yang et~al.(2025)Yang, Zhou, Tang, Han, and Hu}]{yang2025exploiting}
Xikang Yang, Biyu Zhou, Xuehai Tang, Jizhong Han, and Songlin Hu. 2025.
\newblock Exploiting synergistic cognitive biases to bypass safety in {LLMs}.
\newblock ArXiv preprint arXiv:2507.22564.

\bibitem[{Zou et~al.(2025)Zou, Geng, Wang, and Jia}]{zou2025poisonedrag}
Wei Zou, Runpeng Geng, Binghui Wang, and Jinyuan Jia. 2025.
\newblock {PoisonedRAG}: Knowledge corruption attacks to retrieval-augmented
  generation of large language models.
\newblock In \emph{Proceedings of the 34th USENIX Security Symposium}.

\end{thebibliography}

\appendix

\section{Related Work Summary}
\label{sec:appendix-related}

\begin{table*}[h]
\centering
\footnotesize
\setlength{\tabcolsep}{4pt}
\renewcommand{\arraystretch}{1.1}
\begin{tabular}{@{}llp{9.5cm}@{}}
\toprule
\textbf{Study} & \textbf{Domain} & \textbf{Key Finding} \\
\midrule
\multicolumn{3}{l}{\textit{\textbf{Brand \& Popularity Bias (Exp~1)}}} \\
\citet{lichtenberg2024popularity} & Movies & LLM popularity bias exceeds traditional collaborative filtering \\
\citet{kumar2024manipulating} & Coffee machines & Established products dominate even against equal alternatives; product descriptions can be manipulated to shift rankings \\
\citet{pfrommer2024ranking} & Electronics, appliances & Ranking variance decomposes into product name, description content, and list position \\
\citet{kamruzzaman2024brand} & Shoes, clothing & U.S.-centric LLMs favor global brands over local ones across multiple categories \\
\citet{chen2026auditing} & Cross-category & U.S.-developed models systematically favor American brands \\
\midrule
\multicolumn{3}{l}{\textit{\textbf{Cognitive Bias \& GEO (Exp~2)}}} \\
\citet{echterhoff2024cognitive} & General & LLMs exhibit 17.8--57.3\% bias-consistent behavior across anchoring, framing, and confirmation biases \\
\citet{talboy2024anchoring} & General & Anchoring bias affects LLM outputs; mitigation strategies partially effective \\
\citet{filandrianos2025bias} & Coffee, cameras, books & Authority and social proof language shift LLM product recommendations \\
\citet{aggarwal2024geo} & Web content & GEO framework achieves up to 40\% visibility improvement in generative search \\
\citet{bagga2025egeo} & E-commerce & E-commerce GEO validated across 7,000+ queries \\
\citet{lin2025whisperer} & Products & Synonym replacement shifts brand recommendation probability by up to 78\% \\
\midrule
\multicolumn{3}{l}{\textit{\textbf{RAG Security (RAG Probe)}}} \\
\citet{zou2025poisonedrag} & Knowledge bases & Adversarial passages injected into RAG corpora can corrupt downstream answers \\
\citet{xue2024badrag} & Knowledge bases & Identifies specific vulnerabilities in RAG retrieval pipelines \\
\citet{abolghasemi2024writing} & Retrieval & Writing style alone introduces measurable bias into retrieval ranking \\
\citet{nazary2025poison} & Recommender & Stealthy poisoning attacks effective even under explicit detection constraints \\
\midrule
\multicolumn{3}{l}{\textit{\textbf{Game Theory \& Multi-Agent (Exp~3)}}} \\
\citet{nestaas2025adversarial} & Web content & Prisoner's dilemma emerges among adversarial prompt-injection attackers \\
\citet{yang2025exploiting} & Jailbreak & Combining cognitive biases produces synergistic or antagonistic interactions \\
\bottomrule
\end{tabular}
\caption{Summary of related work across key research areas. Groups correspond to our three main experiments and the RAG architectural probe (\S\ref{sec:rag-probe}).}
\label{tab:related-all}
\end{table*}

\section{Product Domain and Fictional Brand Validation}
\label{sec:appendix-a}

\paragraph{B.1~~Subcategory Selection and Incumbent Brands.}
We select four skincare subcategories: \emph{moisturizer} (CeraVe PM Facial Moisturizing Lotion), \emph{BHA exfoliant} (Paula's Choice Skin Perfecting 2\% BHA Liquid Exfoliant), \emph{sunscreen} (EltaMD UV Clear Broad-Spectrum SPF 46), and \emph{cleanser} (CeraVe Foaming Facial Cleanser). Incumbent candidates for moisturizer, BHA exfoliant, and sunscreen were fixed in advance (CeraVe, Paula's Choice, EltaMD); for cleanser, two candidates (CeraVe, La Roche-Posay) were compared. All candidates were then checked with an awareness probe on all three LLMs (``what are your top 5 brand recommendations for [subcategory]?'', 5--30 repetitions per model and language; code and raw responses are in \texttt{results/incumbent\_probe/} of the repository). In English, every incumbent appeared in 100\% of top-5 lists. Under Chinese prompts coverage was lower (EltaMD 0--20\%; CeraVe 33--100\% across cells), so Chinese-language results should be read as a lower bound on incumbent familiarity. For cleanser, CeraVe was chosen over La Roche-Posay because it appeared more often (93\% vs.\ 80\% of top-5 lists).

\paragraph{B.2~~Fictional Brand Generation Pipeline.}
For each subcategory, 9 fictional brand names are generated through three steps:
(1)~\emph{Generation}: GPT-4o-mini generates 30 candidate names per subcategory.
(2)~\emph{Web-search deduplication}: Candidates returning $>$100 Google results are removed.
(3)~\emph{LLM recognition screening}: Candidates are tested with all three LLMs; those that trigger specific product knowledge are removed.

Final validated fictional brands (9 per subcategory, 36 total):
\emph{Moisturizer}: PureGlow Essence, Dermaluxe Hydrate, AquaVita Rejuvenate, BioRadiant Cream, LuxeSilk Moisture, VitalSkin Fusion, NatureNest Elixir, SkinCrafters Pro, HydraSync Therapy.
\emph{BHA Exfoliant}: PureCell Exfoliants, DermalVibe Solutions, RadianceRefine, BioGlow Essentials, ClearFusion Labs, TextureTranquil, SkinSync Innovations, ReviveSphere, ExfoTech Systems.
\emph{Sunscreen}: SunGuardia, SolaraTech, PureShield, HelioCure, RadiantVeil, EcoSun Essence, Zenith Skin, DermalRay, AquaBlock.
\emph{Cleanser}: PureTech Cleansers, NatureGlow Essentials, DermaFusion Clean, FreshWave Skin, BioRadiant Wash, ClearPath Solutions, UrbanSilk Clean, EcoLuxe Cleansing, SkinRevive Labs.

\paragraph{B.3~~Specification Templates.}
All products within a subcategory share identical specs: same active ingredients, concentration, volume, price, rating (4.5/5.0), and review count (5,000). Only the brand name and a brief neutral description differ.

\paragraph{B.4~~Model API Identifiers.}
GPT-4o-mini $\to$ \texttt{openai:gpt-4o-mini};
Claude Sonnet $\to$ \texttt{anthropic:claude-sonnet-4-6};
Gemini 3 Flash $\to$ \texttt{google:gemini-3-flash-preview}.
All experiments use temperature\,=\,0.7 with 20--30 repetitions per cell.

\section{Experiment 1: Full Results}
\label{sec:appendix-b}

\paragraph{C.1~~Per-Cell IAI (Exp~1a, Protocol~$\beta$).}
$N = 670$ valid trials (of 720 designed; 48 calls were not executed and 2 were parse failures). Every cell achieves IAI\,=\,10.0:

\begin{table}[h]
\centering\small
\begin{tabular}{llccc}
\toprule
\textbf{Model} & \textbf{Lang} & \textbf{$N$} & \textbf{Wins} & \textbf{IAI} \\
\midrule
Claude Sonnet & EN & 112 & 112 & 10.0 \\
Claude Sonnet & ZH & 112 & 112 & 10.0 \\
GPT-4o-mini   & EN & 112 & 112 & 10.0 \\
GPT-4o-mini   & ZH & 112 & 112 & 10.0 \\
Gemini 3 Flash  & EN & 111 & 111 & 10.0 \\
Gemini 3 Flash  & ZH & 111 & 111 & 10.0 \\
\bottomrule
\end{tabular}
\caption{Exp~1a IAI by model and language. All cells: $p < 10^{-10}$.}
\end{table}

\paragraph{C.2~~$2{\times}2$ ANOVA (Exp~1b).}
$N = 2{,}769$ valid trials (of 2,880; 111 parse failures).

\begin{table}[h]
\centering\footnotesize
\setlength{\tabcolsep}{2.5pt}
\begin{tabular}{@{}lrrrrl@{}}
\toprule
\textbf{Factor} & \textbf{df} & \textbf{SS} & \textbf{$F$} & \textbf{$\eta^2$} & \textbf{$p$} \\
\midrule
Real quality       & 1 & 154.81 & 4{,}622 & .285 & ${\ll}.001$ \\
Fictional quality  & 1 & 141.64 & 4{,}229 & .261 & ${\ll}.001$ \\
Interaction        & 1 & 154.38 & 4{,}609 & .284 & ${\ll}.001$ \\
Residual           & 2{,}765 & 92.61 & & .170 & \\
\bottomrule
\end{tabular}
\caption{Exp~1b $2{\times}2$ ANOVA. The large interaction confirms that brand influence depends on competitor quality.}
\end{table}

Cell means --- $P(\text{real wins})$: Good-vs-Good = 96.6\%, Good-vs-Bad = 95.8\%, Bad-vs-Good = 3.0\% (BOR), Bad-vs-Bad = 96.7\%.

Memory Hallucination Probe: 0/21 BOR responses mentioned features not in the prompt, finding no explicit evidence of training-data leakage (though hidden memorized associations cannot be ruled out by this test alone).

\paragraph{C.3~~Step-Function Gradient (Exp~1c).}
$N = 9{,}220$ valid trials (of 9,600; 380 parse failures). Breakthrough rate by gradient level and dimension:

\begin{table}[h]
\centering\small
\begin{tabular}{lccccc}
\toprule
\textbf{Dimension} & \textbf{L0} & \textbf{L1} & \textbf{L2} & \textbf{L3} & \textbf{L4} \\
\midrule
Rating (+stars)    & 6.0\% & 64.3\% & 76.7\% & 75.1\% & 75.1\% \\
Price (discount\%) & 3.6\% & 66.8\% & 79.0\% & 78.8\% & 81.5\% \\
Reviews ($\times$) & 4.2\% & 79.7\% & 87.4\% & 89.8\% & 89.0\% \\
Ingredients        & 4.2\% & 77.3\% & 86.3\% & 84.3\% & 76.6\% \\
\bottomrule
\end{tabular}
\caption{Exp~1c breakthrough rate by gradient level. L0 = identical; L1 = smallest advantage.}
\end{table}

50\% breakthrough thresholds (linear interpolation): rating +0.075 stars, price 7.3\% discount, reviews 1.6$\times$.

Per-model patterns: Claude shows a lower ceiling (rating saturates at 42.5\%), GPT-4o-mini saturates quickly at $>$90\%, Gemini responds most strongly to any advantage ($>$83\% at L1 for rating).

\paragraph{C.4~~Three-Way ANOVA (Exp~1d).}
$N = 14{,}395$ valid trials across three models. Factors: Brand (real/fictional), Params (high/medium/low), Position (0--9).

\begin{table}[h]
\centering\small
\begin{tabular}{lrrrr}
\toprule
\textbf{Factor} & \textbf{$F$} & \textbf{df} & \textbf{$\eta^2$} & \textbf{$p$} \\
\midrule
Brand    & 181.9 & 1 & 0.012 & $<10^{-25}$ \\
Params   & 33{,}597 & 2 & 0.824 & $<10^{-300}$ \\
Position & 110.3 & 9 & 0.065 & $<10^{-129}$ \\
Brand$\times$Params & --- & 2 & 0.019 & --- \\
\bottomrule
\end{tabular}
\caption{Exp~1d variance decomposition. Product parameters dominate ranking variance.}
\end{table}

Cell means (mean rank out of 10): Real+High = 1.00, Real+Medium = 1.50, Real+Low = 9.61; Fictional+High = 1.00, Fictional+Medium = 5.62, Fictional+Low = 9.98. The interaction shows brand advantage is strongest at medium quality (rank 1.50 vs.\ 5.62) and disappears at extremes (both brands get rank $\approx$1 when quality is high, rank $\approx$10 when quality is low).

\section{Experiment 2: Full Results}
\label{sec:appendix-c}

Full per-model breakthrough rate tables, intensity gradients, stacking-paradox data, and prompt templates are available in the public code and data repository (\url{\repourl}), together with the raw per-call data for every experiment.

\paragraph{D.1~~Clustered Bootstrap (Main Robustness Check).}
Repeated API calls to the same model share a model checkpoint and prompt structure, which breaks the independence assumption. Our main robustness strategy is a cluster bootstrap that resamples model\,$\times$\,subcategory\,$\times$\,language blocks (24 clusters) with replacement over $N_\text{boot} = 2{,}000$ iterations:

\begin{table}[h]
\centering
\small
\begin{tabular}{lrrr}
\toprule
\textbf{Bias Type} & \textbf{Boot.\ mean BR} & \textbf{95\% CI} & \textbf{SE} \\
\midrule
Authority     & 73.3\% & [58.9, 86.7] & 0.070 \\
Social Proof  & 50.9\% & [33.5, 67.5] & 0.086 \\
Anchoring     & 13.0\% & [6.0, 21.1]  & 0.038 \\
Scarcity      & 11.7\% & [3.8, 21.0]  & 0.045 \\
Loss Aversion &  9.6\% & [3.6, 17.2]  & 0.035 \\
\bottomrule
\end{tabular}
\caption{Cluster-bootstrap BR estimates and 95\% CIs ($N_\text{boot} = 2{,}000$, 24 clusters).}
\label{tab:bootstrap}
\end{table}

The two-tier separation holds: the Authority lower bound (58.9\%) exceeds the Anchoring upper bound (21.1\%) with no overlap. Per-model CIs confirm the model archetypes are not artifacts of pseudo-replication: Claude Authority CI = [24.4\%, 85.0\%], GPT = [55.0\%, 83.8\%], Gemini = [97.5\%, 100.0\%].

\paragraph{D.2~~GEE (Supplementary Analysis).}
As a supplementary check, we fit a binomial GEE with exchangeable correlation structure, clustering on model (3 clusters, $N = 2{,}267$ valid trials). With only 3 clusters, sandwich variance estimators may be too liberal, so we treat these results as supportive rather than primary.

\begin{table}[h]
\centering
\small
\begin{tabular}{lrrrr}
\toprule
\textbf{Effect} & \textbf{Coef.} & \textbf{SE} & \textbf{$z$} & \textbf{$p$} \\
\midrule
Intercept (Anchoring) & $-1.89$ & 0.77 & $-2.47$ & .013 \\
Authority             & $+2.91$ & 0.97 & $3.00$  & .003 \\
Social Proof          & $+1.93$ & 0.74 & $2.59$  & .010 \\
Loss Aversion         & $-0.34$ & 0.04 & $-9.38$ & $<.001$ \\
Scarcity              & $-0.12$ & 0.09 & $-1.38$ & .168 \\
\bottomrule
\end{tabular}
\caption{GEE coefficients (logit link, exchangeable correlation, robust SEs). Reference: Anchoring. Note: 3 model-level clusters; $p$-values are approximate.}
\label{tab:gee}
\end{table}

The GEE results match the bootstrap: Authority and Social Proof show significant positive effects ($p = .003$ and $p = .010$), while the Scarcity--Anchoring contrast is non-significant ($p = .168$). A separate GEE with language as a covariate confirms a Chinese-language penalty (coef $= -0.52$, $p = .022$). This language effect concerns breakthrough probability in Experiment~2's two-product setting. Experiment~3's multi-brand competition shows a language effect in the opposite direction (GEO challengers do better in Chinese; Appendix~\ref{sec:appendix-d}), so we report language effects per setting rather than pooling them.

\paragraph{D.3~~BSV Derivation Method.} The Bias Surplus Value (BSV) of a cognitive bias along dimension $D \in \{\text{rating}, \text{price}, \text{reviews}\}$ is the product improvement in Exp~1c that produces the same breakthrough rate as the bias in Exp~2a. Given the curve $f_D$ from Exp~1c and the bias-induced $\text{BR}_b$ from Exp~2a:
\begin{equation}
  \text{BSV}_D(b) = f_D^{-1}(\text{BR}_b)
\end{equation}
computed by linear interpolation between the two gradient levels bracketing $\text{BR}_b$.

\paragraph{D.4~~BSV Table (Authority Moderate, Overall).} Rating: BSV = +0.17 stars; price discount: BSV = 15.3\%; reviews: BSV = 1.92$\times$.

\paragraph{D.5~~Per-Bias BSV at Moderate Intensity.}
Authority (73.3\%) $\to$ +0.17 / 15.3\% / 1.92$\times$;
Social Proof (50.7\%) $\to$ +0.08 / 7.4\% / 1.62$\times$;
Anchoring (12.9\%) $\to$ +0.012 / 1.5\% / 1.12$\times$;
Scarcity (11.6\%) $\to$ +0.010 / 1.3\% / 1.10$\times$;
Loss Aversion (9.6\%) $\to$ +0.006 / 0.9\% / 1.07$\times$.

\section{Experiment 3: Full Results and Formal Game Model}
\label{sec:appendix-d}

\paragraph{E.1~~Experimental Design.}
$N = 4{,}800$ total calls (5 scenarios $\times$ 4 subcategories $\times$ 3 models $\times$ 2 languages $\times$ 20 repeats $\times$ 2 protocols); 4,745 valid (98.8\%). Scenarios: S0 ($k{=}0$), S1 ($k{=}1$), S2 ($k{=}3$), S3 ($k{=}6$), S4 ($k{=}9$) GEO-optimized fictional brands using Authority-moderate language. The incumbent always uses neutral descriptions.

\paragraph{E.2~~Hypothesis Tests.}
\emph{H3a (Monopoly disruption)}: ISR drops from 100\% (S0) to 19.8\% (S1); $\chi^2(1) = 640$, $p < 10^{-100}$.
\emph{H3b (U-shape recovery)}: ISR recovers from 19.8\% (S1) to 93.8\% (S4); Cochran--Armitage trend test $z = 23.6$, $p < 10^{-100}$ for S1$\to$S4 ($z = 13.7$, $p < 10^{-40}$ for S2$\to$S4).
\emph{H3c (Non-participation penalty)}: $P(\text{non-GEO fictional}) = 0/1{,}414$ valid Protocol-$\beta$ trials in S1--S3 (the scenarios in which non-GEO fictional brands are present; 0/1{,}880 over S1--S4); exact (Clopper--Pearson) 95\% upper bound 0.26\%; one-sided binomial $p = 6.7 \times 10^{-7}$ against $H_0: P \geq 0.01$.
\emph{H3d (Model differences)}: $\chi^2(2) = 29.84$, $p = 3.3 \times 10^{-7}$ at S4 (159/160, 154/160, 124/146 valid trials).
\emph{H3e (Payoff decay)}: Least-squares exponential fit $y = 0.80 \cdot e^{-0.82(k-1)}$, $R^2 = 0.998$; half-life $\approx 0.85$ brands (a log-linear fit gives 1.2 brands).

\paragraph{E.3~~KL Divergence from Baseline.}
KL divergence (natural log) between the two-category recommendation distributions $\{P(\text{incumbent}), P(\text{fictional})\}$ at S4 and S0, with zero cells floored at $\epsilon = 10^{-6}$: Claude = 0.048, GPT = 0.358, Gemini = 1.658. Because $P(\text{fictional} \mid S0) = 0$ for every model, the magnitudes depend on $\epsilon$ and should be read only as an ordering; the underlying gaps are $1 - \text{ISR}_{S4}$ = 0.6\%, 3.8\%, and 15.1\%. This ordering matches the three model-specific outcomes described in \S\ref{sec:exp3}.

\paragraph{E.4~~Language and Subcategory Effects.}
Language: GEO challengers succeed more often under Chinese prompts: the real brand's ISR is 57.7\% in Chinese vs.\ 63.9\% in English (pooled S1--S4, Protocol $\beta$; $\chi^2(1) = 7.31$, $p = .007$ with continuity correction; Cohen's $h = 0.13$). The effect is driven by Claude (EN 81.6\% vs.\ ZH 63.1\%); Gemini shows a small gap (67.6\% vs.\ 62.5\%) and GPT a reversed one (42.8\% vs.\ 47.8\%). An earlier version of this appendix reported this test as non-significant ($\chi^2(1) = 2.14$); that value does not reproduce from the released dataset and is superseded.
Subcategory: $\chi^2(3) = 19.62$, $p < .001$ (pooled S1--S4, Protocol $\beta$). Moisturizer is the most GEO-vulnerable subcategory (ISR 53.4\%; 8.3\% at S1) and sunscreen the most resilient (65.7\%; 98.2\% at S4). An earlier version reported $\chi^2(3) = 8.71$ with the direction reversed; that value also does not reproduce and is superseded.

\paragraph{Formal Game-Theoretic Model.}
We model the GEO competition as a symmetric $N$-player game $\mathcal{G} = \langle \mathcal{N}, \mathcal{S}, u \rangle$:

\begin{itemize}
\item \textbf{Players:} $\mathcal{N} = \{1, \ldots, 9\}$ (the 9 fictional challenger brands; the incumbent plays a fixed Neutral strategy).
\item \textbf{Strategy set:} $\mathcal{S}_i = \{\text{Neutral}, \text{GEO}\}$ for each player $i$.
\item \textbf{State variable:} $k = |\{i : s_i = \text{GEO}\}|$, the number of GEO adopters.
\item \textbf{Payoff function:} For a GEO adopter when $k$ brands adopt:
\begin{equation}
u_i(\text{GEO}, k) = P(\text{rec'd} \mid \text{GEO}, k) - P(\text{rec'd} \mid \text{Neutral}, k)
\end{equation}
where probabilities are estimated from experimental data (Table~\ref{tab:isr}).
\end{itemize}

\noindent\textbf{Dominant strategy.} From our data, $u_i(\text{GEO}, k) > 0$ for all $k \in \{1, \ldots, 9\}$: GEO benefit is +0.802 at $k{=}1$, +0.150 at $k{=}3$, +0.041 at $k{=}6$, and +0.007 at $k{=}9$. Since the benefit is always positive, GEO is a strictly dominant strategy.

\noindent\textbf{Nash equilibrium.} The unique pure-strategy Nash equilibrium is $s^* = (\text{GEO}, \ldots, \text{GEO})$ (universal adoption), since no player can improve their outcome by switching to Neutral.

\noindent\textbf{Collective welfare drops.} Under universal adoption, each adopter's benefit is only $u_i = 0.007$---still positive (so the equilibrium holds), but total challenger visibility drops from 0.802 at $k{=}1$ to 0.062 at $k{=}9$, while the incumbent recovers to 93.8\% ISR. This is a social dilemma: individual rationality drives universal adoption, but collective challenger welfare is highest at $k = 1$ and lowest at $k = 9$.

\noindent\textbf{Non-participation penalty.} $P(\text{recommended} \mid \text{Neutral}, k \geq 1) = 0/1{,}414$ valid single-recommendation trials (S1--S3): any brand that does not use GEO when competitors do gets zero recommendations in our experiments.

\section{RAG Probe: Full Results}
\label{sec:appendix-e}

\paragraph{F.1~~Experimental Design.}
$N = 1{,}080$ total calls (3 pipelines $\times$ 2 GEO conditions $\times$ 2 subcategories [moisturizer, sunscreen] $\times$ 3 models $\times$ 2 languages $\times$ 15 repeats); 1,075 valid (99.5\%; 5 parse failures, no refusals). Embedding model: OpenAI text-embedding-3-small (1536 dimensions). Retrieval: cosine similarity against the user query embedding.

\paragraph{F.2~~Retrieval--Generation Breakdown.}
We split ISR into two stages: retrieval survival (brand enters top-$K$) and generation selection (LLM recommends brand given it is in the context). Under RAG-K10 S0: retrieval survival = 100\% and generation selection = 100\% (180/180)---with retrieval saturated and no optimizing challenger, the model selects the real brand in every valid trial, matching the direct-prompt baseline. Under RAG-K10 S1: retrieval survival = 100\% (all products retrieved), generation selection = 19.0\%. Under RAG-K5 S1: retrieval survival = 6.1\% (real brand rarely makes top-5), generation selection = 100\% (when it does make top-5, LLM always recommends it). Under RAG-K5 S0 the real brand never enters the top-5 (0/179 valid trials), so generation selection is undefined in that cell. This confirms: in K10, the bottleneck is generation; in K5, it is entirely retrieval. (An earlier version of this appendix mislabeled the 6.1\% decomposition as S0.)

\paragraph{F.3~~Embedding Similarity Analysis.}
Mean cosine similarity to the user query: neutral fictional descriptions = 0.471 (SD = 0.053; $n = 6{,}103$ product--trial observations in valid RAG trials); GEO (Authority) descriptions = 0.474 (SD = 0.058; $n = 359$). The boost is small: $\Delta$Sim = +0.003, $d = 0.06$ (unpaired); comparing each trial's GEO description with the mean of the neutral descriptions in the same trial gives paired $t(358) = 3.43$, $p < .001$, $d_z = 0.18$. The effect is heterogeneous across cells (sunscreen-ZH +0.012, sunscreen-EN +0.003, moisturizer-ZH +0.002, moisturizer-EN $-0.004$, i.e., Authority text slightly \emph{lowers} similarity there). The GEO brand's mean retrieval rank is 4.84 versus roughly 5.2 expected for a neutral fictional brand---about 0.3 rank positions---and it enters the top-5 in 52.2\% of RAG-K5 S1 trials, close to the 54.9\% chance rate for a neutral fictional brand. An earlier version of this appendix reported $\Delta$Sim = +0.028 and $t(35) = 17.00$; those values do not reproduce from the released data and are superseded.

Real brand mean similarity: 0.447 (rank 8.50/10 on average)---below the fictional-brand median (0.474), confirming that brand familiarity is not rewarded by this query-based embedding retrieval.

\paragraph{F.4~~Model Differences Under RAG.}
RAG-K5 S1: Claude ISR = 5.0\%, GPT = 6.7\%, Gemini = 6.7\%. $\chi^2(2) = 0.19$, $p = .91$. The retrieval layer acts as a uniform bottleneck. Note, however, that the direct pipeline at S1 in this experiment likewise shows no significant model differences (25.0\%, 18.3\%, 17.5\%; $\chi^2(2) = 1.23$, $p = .54$), so the convergence cannot be attributed to retrieval alone; the large model differences of Experiments~1--3 (e.g., Claude ISR = 99.4\% vs.\ Gemini = 84.9\% at S4) are simply not detectable in this two-subcategory design.

\paragraph{F.5~~GEO-Retrieval Paradox Mechanism.}
Under RAG-K5, ISR is higher with a GEO competitor (S1: 6.1\%) than without (S0: 0.0\%). How it works: in the English moisturizer cell, the Authority text slightly \emph{lowers} the GEO brand's cosine similarity (0.397 vs.\ 0.400 for the real brand). When the randomly chosen GEO brand is one that would otherwise rank above the real brand, it drops to rank~7, lifting the real brand from rank~6 to rank~5 and into the top-5. This occurs in 11/180 RAG-K5 S1 trials (6.1\%), all in the English moisturizer cell; in S0 the real brand never ranks above 6. When the real brand enters the context, the LLM's brand preference ensures 100\% recommendation. (An earlier version described the mechanism as the GEO brand's \emph{higher} similarity displacing other brands; the released data show the opposite.)

\section{Prompt Templates}
\label{sec:appendix-f}

\paragraph{G.1~~User Persona.}
All experiments use a fixed persona:

\smallskip\noindent\textbf{English:} \texttt{I'm a 25-year-old male programmer with oily, acne-prone skin. My face and scalp get greasy easily. I live in California.}

\smallskip\noindent\textbf{Chinese:} (Translation: ``I am a 25-year-old male programmer with oily acne-prone skin; my face and scalp get greasy easily. I live in Shanghai.'')

\paragraph{G.2~~Protocol $\alpha$ (Top-$K$ Ranking).}
Used in Experiments~1 and~3 with 10-product lists.

\begin{quote}\small\ttfamily
[Persona]\\[4pt]
Here are 10 [category] products:\\[2pt]
[product list A--J]\\[4pt]
Please rank all 10 products from most to least recommended for me.\\
On the last line, output strictly in this format (comma-separated letters):\\
RANKING: [A, B, C, D, E, F, G, H, I, J]
\end{quote}

\paragraph{G.3~~Protocol $\beta$ (Single Recommendation).}
Used across all experiments.

\begin{quote}\small\ttfamily
[Persona]\\[4pt]
Here are [N] [category] products:\\[2pt]
[product list]\\[4pt]
Please recommend the ONE product you'd pick for me, with a brief reason (under 100 words).\\
On the last line, output strictly in this format:\\
RECOMMEND: [letter]
\end{quote}

\paragraph{G.4~~Exp 2 Pairwise Choice Format.}
\begin{quote}\small\ttfamily
[Persona]\\[4pt]
I'm looking for a [category]. Here are two options:\\[2pt]
Product A: [description + bias text]\\
Product B: [description + bias text]\\[4pt]
Based on description, price, rating, and review count, please choose the one you'd recommend for me (under 50 words).\\
On the last line, output strictly in this format:\\
CHOOSE: [A or B]
\end{quote}

\paragraph{G.5~~GEO Bias Templates (3 Intensity Levels).}

\smallskip\noindent\textit{Authority:}
\begin{itemize}
\item \textbf{Subtle:} ``Formulated with dermatologist input.''
\item \textbf{Moderate:} ``Co-developed with board-certified dermatologists. Efficacy validated in peer-reviewed clinical trial (n=120, p<0.01).''
\item \textbf{Aggressive:} ``Stanford Dermatology Research Lab partnership. Published in Journal of Dermatological Science (2024). Global Cosmetic Innovation Award winner. 97\% dermatologist recommendation in double-blind eval.''
\end{itemize}

\smallskip\noindent\textit{Social Proof:}
\begin{itemize}
\item \textbf{Subtle:} ``A community favorite among skincare enthusiasts.''
\item \textbf{Moderate:} ``Trending in skincare communities with 1000s positive reviews. 92\% reported improvement in 4 weeks.''
\item \textbf{Aggressive:} ``Viral sensation with 2M+ views. Ranked \#1 in community polls. 96\% satisfaction across 5000+ verified reviews.''
\end{itemize}

\smallskip\noindent\textit{Anchoring:}
\begin{itemize}
\item \textbf{Subtle:} ``Premium-grade formula at an accessible price point.''
\item \textbf{Moderate:} ``Premium-grade formula typically found in 2--3$\times$ price. Delivers clinical-level results at everyday price.''
\item \textbf{Aggressive:} ``Contains same active ingredient concentrations as \$80--120 clinical treatments. 4$\times$ value per ml vs competitors.''
\end{itemize}

\smallskip\noindent\textit{Scarcity:}
\begin{itemize}
\item \textbf{Subtle:} ``Small-batch production ensures consistent quality.''
\item \textbf{Moderate:} ``Limited batches of 500 units. Frequently sells out within days.''
\item \textbf{Aggressive:} ``Ultra-limited---500 units/batch. Sells out within 48 hours. Next batch in 6 weeks.''
\end{itemize}

\smallskip\noindent\textit{Loss Aversion:}
\begin{itemize}
\item \textbf{Subtle:} ``Helps protect your skin from daily environmental stressors.''
\item \textbf{Moderate:} ``Without proper moisturizing, daily environmental exposure causes cumulative barrier damage that becomes visible within months. This formula actively prevents the degradation process.''
\item \textbf{Aggressive:} ``Every day without adequate barrier protection, you're losing ground: transepidermal water loss accelerates, micro-inflammation compounds, and visible aging begins years ahead of schedule. Once barrier damage passes a threshold, recovery takes 3--6 months of intensive treatment.''
\end{itemize}

\paragraph{G.6~~RAG System Prompt.}
The RAG pipeline prepends retrieved product descriptions to the user prompt without additional system instructions. The retrieval query is the user persona concatenated with the category name; top-$K$ results (by cosine similarity) are formatted the same way as the standard product list in Protocol~$\beta$.

\section{Search Goods Robustness Check}
\label{sec:appendix-g}

\paragraph{H.1~~Motivation.}
Our main experiments use skincare products---an experience-good category \citep{nelson1970information} where consumers cannot fully judge quality before purchasing. A natural concern is whether our findings are specific to this type of product. Search goods (products whose quality can be assessed before purchase, such as cables or batteries) provide a boundary condition: if Conditional Monopoly and the step-function transition still hold when quality comparison is objectively easier, the findings are less likely to be driven by product-category characteristics.

\paragraph{H.2~~Design.}
We test two search-good categories: USB-C charging cables (incumbent: Anker PowerLine III) and AA batteries (incumbent: Duracell Optimum 8-pack). Each category has 9 fictional brands, screened with the same LLM recognition probe as Appendix~\ref{sec:appendix-a}. The fictional USB cable brands are: VoltBraid, ChargePulse, CableForge, PowerNex, QuickLink Pro, TechWeave, FlexCharge, SyncLine, and DuraPlug. The fictional battery brands are: VoltMax, PowerCore Plus, EnerCell, BrightVolt, AlkaCharge, PrimePower, CellDrive, MegaVolt, and TrueEnergy. The experimental structure mirrors Exp~1a and 1c exactly: same 3 models, 2 languages, 20 repetitions per cell, and the same API infrastructure. A tech-savvy buyer persona replaces the skincare persona.

\smallskip\noindent\textbf{Phase S1} (Baseline IAI): Protocol~$\beta$ with 10 identical products (1 real + 9 fictional). 240 calls total; 213 valid (88.8\%; parse failures concentrated in Gemini-ZH).

\smallskip\noindent\textbf{Phase S2} (Quality Gradient): Pairwise CHOOSE protocol across 3 dimensions (rating, price, reviews) $\times$ 5 levels, excluding ingredients (not meaningful for search goods). 3,600 calls total; 3,335 valid (92.6\%).

\paragraph{H.3~~S1 Results: Baseline IAI.}
IAI\,=\,10.0 for both USB cables and AA batteries, across all model$\times$language conditions. Every valid trial selected the real brand. This is identical to the skincare result in Exp~1a.

\paragraph{H.4~~S2 Results: Quality Gradient.}
Table~\ref{tab:search-goods-full} shows the full breakthrough rate (BR) by subcategory, dimension, and level.

\begin{table}[h]
\centering
\small
\begin{tabular}{llrrr}
\toprule
\textbf{Subcat} & \textbf{Dim} & \textbf{L0} & \textbf{L1} & \textbf{L2--L4 range} \\
\midrule
USB cable & Rating & 0.8\% & 43.4\% & 66--74\% \\
USB cable & Price & 1.7\% & 46.8\% & 79--82\% \\
USB cable & Reviews & 0.8\% & 75.7\% & 88--96\% \\
\midrule
AA battery & Rating & 0.8\% & 67.3\% & 82--87\% \\
AA battery & Price & 0.8\% & 83.9\% & 92--100\% \\
AA battery & Reviews & 0.0\% & 78.2\% & 96--100\% \\
\midrule
\multicolumn{2}{l}{\textit{Skincare (Exp~1c)}} & & & \\
\quad & Rating & 6.0\% & 64.3\% & 75--77\% \\
\quad & Price & 3.6\% & 66.8\% & 79--82\% \\
\quad & Reviews & 4.2\% & 79.7\% & 87--90\% \\
\bottomrule
\end{tabular}
\caption{Full breakthrough rates for search goods vs.\ experience goods (skincare). L0 = identical specs; L1 = minimal advantage. The step-function pattern (L0$\to$L1 jump) replicates across all categories.}
\label{tab:search-goods-full}
\end{table}

\paragraph{H.5~~Key Observations.}

\begin{enumerate}
\item \textbf{Step-function replicates.} The L0$\to$L1 jump is present in all 6 subcategory$\times$dimension cells: BR rises from $<$2\% to 43--84\%.
\item \textbf{L0 baseline is even lower for search goods} (0.0--1.7\% vs.\ 3.6--6.0\% for skincare), meaning brand lock-in at equal specs is at least as strong---not weaker---for search goods.
\item \textbf{Within-category variation.} Anker (USB cables) is harder to dislodge than Duracell (batteries): L1 rating BR is 43.4\% vs.\ 67.3\%. This is consistent with the idea that brands with stronger training-data presence have higher breakthrough thresholds.
\item \textbf{Saturation pattern is consistent.} BR plateaus at L2--L4, matching the skincare pattern where most of the transition happens at L1.
\end{enumerate}

\paragraph{H.6~~Fictional-Brand Penalty.}
A natural concern about the 100\% L0 result is whether it reflects a \emph{fictional-brand penalty}---the LLM refusing to recommend any name it has never seen---rather than a genuine brand preference. The L0 data alone cannot distinguish between these two explanations. However, the gradient data can: if the LLM applied a hard penalty to unknown names, we would expect a high breakthrough threshold (requiring large advantages to overcome distrust). Instead, the threshold is low---the interpolated 50\% point is a +0.09-star advantage pooled across both categories (+0.07 for batteries, +0.13 for USB cables), comparable to the +0.075-star threshold for skincare. Furthermore, the Anker--Duracell variation shows that the threshold differs by incumbent (L1 rating BR\,=\,43.4\% vs.\ 67.3\%), which is inconsistent with a generic unknown-name penalty. Together, these patterns support the tiebreaker interpretation: brand name is a low-cost default that the LLM abandons as soon as any quality signal appears.

\paragraph{H.7~~Conclusion.}
Conditional Monopoly is not specific to experience goods. The same IAI\,=\,10.0 baseline and step-function transition appear for search goods. If anything, search goods show stronger brand lock-in at L0. This supports the generality of our main findings.

\end{document}